\documentclass[preprint,5p,times,twocolumn,authoryear]{elsarticle}

\usepackage{amssymb}
\usepackage{amsmath}
\usepackage{lipsum}
\usepackage[colorlinks,citecolor=red,urlcolor=blue,bookmarks=false,hypertexnames=true]{hyperref}
\usepackage{xcolor} 
\usepackage{soul} 
\usepackage{csquotes}
\usepackage{float} 

\setcitestyle{authoryear,open={(},close={)}}

\journal{arXiv}

\begin{document}

\begin{frontmatter}

\title{Investigating the Segment Anything Foundation Model for Mapping Smallholder Agriculture Field Boundaries Without Training Labels}

\author{Pratyush Tripathy}
\ead{ptripathy@ucsb.edu}

\author{Kathy Baylis}
\ead{baylis@ucsb.edu}

\author{Kyle Wu}
\ead{klw@ucsb.edu}

\author{Jyles Watson}
\ead{jyleswatson@ucsb.edu}

\author{Ruizhe Jiang}
\ead{ruizhejiang@ucsb.edu}

\address{Department of Geography, University of California, Santa Barbara, CA, 93106 USA}

\begin{abstract}
Accurate mapping of agricultural field boundaries is crucial for enhancing outcomes like precision agriculture, crop monitoring, and yield estimation. However, extracting these boundaries from satellite images is challenging, especially for smallholder farms and data-scarce environments. This study explores the Segment Anything Model (SAM) to delineate agricultural field boundaries in Bihar, India, using 2-meter resolution SkySat imagery without additional training. We evaluate SAM's performance across three model checkpoints, various input sizes, multi-date satellite images, and edge-enhanced imagery. Our results show that SAM correctly identifies about 58\% of field boundaries, comparable to other approaches requiring extensive training data. Using different input image sizes improves accuracy, with the most significant improvement observed when using multi-date satellite images. This work establishes proof of concept for using SAM and maximizing its potential in agricultural field boundary mapping. Our work highlights SAM's potential in delineating agriculture field boundary in training-data scarce settings to enable a wide range of agriculture related analysis.
\end{abstract}



\end{frontmatter}



\section{Introduction}
\label{introduction}

Images captured by remote sensing satellites have been widely used for agriculture monitoring, including crop type, tillage mapping, harvest date, yield estimation, and crop health \citep{weiss2020}. However, these crop maps are more meaningful (and often more accurate) at the field level rather than the pixel level \citep{north2018}. In addition to improving agriculture-related statistics and decision-making, agriculture field boundaries also contribute to analyses related to air pollution, groundwater, and socioeconomic conditions \citep{estes2021, garcia-pedrero_outlining_2018, campbell_introduction_2011}. Despite their critical role, obtaining up-to-date information on agricultural field boundaries remains challenging, particularly in low- and middle-income regions such as South Asia and Africa, which have the largest number of farmers and food-insecure populations \citep{singh2002smallholder, gandhi_food_2014}. Access to accurate field boundary data in these regions is crucial for maximizing yields, promoting sustainable agricultural practices, and ensuring global food security \citep{gafurov_automated_2023}.

Despite the dire need, traditional methods of generating agricultural field boundaries through surveys and manual creation are neither efficient nor cost-effective across large spatial and temporal scales \citep{taravat_advanced_2021, garcia-pedrero_outlining_2018}. A more effective solution to generate agriculture boundary dataset is using high-resolution satellite images \citep{wang_agricultural_2022}. Previous studies extracting agricultural field boundaries from satellite data can be categorized into two approaches: computer vision-based segmentation techniques \citep{DEBATS2016210} and more recent Deep Learning (DL) methods, which have achieved unprecedented accuracy and generalizability \citep{waldner_deep_2020}.

However, detecting smallholder farm boundaries from satellite images using DL faces two major challenges. First, DL models require large amounts of training data (several thousand samples), which is often unavailable in low- and middle-income countries \citep{PERIKAMANA2024171}. Second, most studies focus on regions with large agricultural fields (e.g., 2—10 ha). In contrast, agriculturally active regions like Bangladesh, Nigeria, and the Indo-Gangetic plain are dominated by smallholder farms ($<$0.6 ha), making automated field mapping particularly challenging \citep{YANG2020100413, zhang2020}.

While large amounts of training data for DL models remain scarce, recent studies have addressed this using smaller datasets \citep{wang2022, mei2022}. \cite{wang2022} demonstrate a transfer learning approach, where a model trained on data from France is fine-tuned using limited training samples from India. \cite{mei2022} present a two-step training approach where the model is initially trained on a small amount of data to produce preliminary predictions, which are then refined using GIS techniques and used as training labels for further training. These methods reduce the amount of ground truth field boundary data needed but still require field boundary data. A solution to this problem in regions that do not have any training data at all may be utilizing generalist Artificial Intelligence (AI) models, such as Foundation Models (FMs).

\subsection{Foundation models and geospatial applications}
FMs are trained on vast amounts of diverse data and can be applied to specific tasks with minimal (few-shot learning) or no additional training (zero-shot learning). The rise in computational resources and training data has increased the use of FMs in the geospatial domain \citep{wang2024}. Examples of geospatial FMs include Autonomous GIS \citep{li_2023}, GeoGPT \citep{zhang_geogpt_2023}, MapGPT \citep{fernandez_core_2023}, MOSAIKS \citep{rolf_2021}, and Prithvi \citep{jakubik_2023}. While the availability of FMs is expanding, evaluating their reliability for various geospatial tasks is crucial. This paper investigates the suitability of one such publicly available FM, the Segment Anything Model (SAM), for agricultural field boundary detection \citep{sam2023}.

\subsection{Segment Anything Model}
SAM is a Vision Transformer (ViT) based model with an image encoder, a flexible prompt encoder, and a fast mask decoder, trained on over 1 billion masks in 11 million images \citep{sam2023}. Although SAM is not specifically trained on satellite images or for agricultural field boundary detection, its zero-shot learning and extensive training data enable it to segment various images without additional training. However, its performance on remote sensing images, particularly for agricultural field boundary delineation, is unknown. Therefore, this paper investigates SAM's efficacy in delineating smallholder agricultural field boundaries from satellite images without additional training. While SAM can be refined to work with satellite datasets or specifically segment agricultural field boundaries using its promptable feature, this paper limits itself to evaluating the model's performance without any further training or fine-tuning.

\subsection{SAM for agriculture field boundary delineation}
The application of SAM to delineate agricultural field boundaries (and geospatial tasks in general) involves several critical decisions that this paper explores. First, three different checkpoints of SAM are available—ViT-b, ViT-h, and ViT-l—representing saved states at different training points. It is unknown which checkpoint is ideal or if they are all consistent for segmenting agricultural fields. Second, SAM can segment images of any input size, which is advantageous for general images but challenging for remote sensing images since objects on the ground have a size component, and the ideal input image size is unknown. Third, the reflectance of sowed agriculture fields changes over the crop growth season, suggesting that using satellite data from multiple dates might be beneficial, but whether or not SAM can leverage this is unknown. We provide empirical investigation into whether using images from different dates improves segmentation results, especially for FMs and agricultural boundary delineation. Fourth, studies have shown that feature engineering techniques, such as enhancing image contrast, may not necessarily improve segmentation accuracy for agricultural field delineations \citep{mei2022}. Therefore, there is a lack of consensus on whether edge enhancement techniques improve segmenting features from satellite images. To explore these aspects, this paper investigates the following questions regarding the utility of SAM for agricultural field boundary delineation:

\begin{enumerate}
    \item Do different checkpoints of SAM (ViT-b, ViT-h, and ViT-l) produce consistent results, or does one perform better? Can the accuracy be improved by combining predictions from these checkpoints?
    \item Does the dimension of the input image impact segmentation accuracy? Can the accuracy be improved by combining predictions from images of different sizes?
    \item Are the predictions consistent when using satellite images from different dates? Can the accuracy be improved by combining predictions from different dates?
    \item Does enhancing the edge of the image improve segmentation accuracy?
    \item How well can SAM perform in segmenting smallholder agricultural field boundaries by combining outputs from different combinations (checkpoints, input image dimensions, image acquisition dates) without additional training?
\end{enumerate}

\section{Data and methods}
\subsection{Satellite data}
To investigate the utility of SAM for agricultural field boundary delineation, we use 2 m resolution SkySat orthorectified multispectral satellite data provided by Planet Labs Inc., available on Google Earth Engine (GEE) \citep{planetskysat}. While other studies often use 0.5 m resolution data for similar tasks, higher resolution data is difficult to obtain and more computationally burdensome. Our use of 2 m data balances resolution and computational efficiency, making it practical for large-scale applications. Details of the satellite data used are provided in Section \ref{methods:different_dates_image}.

\begin{figure*}[t]
  \centering
  \includegraphics[width=0.6\textwidth]{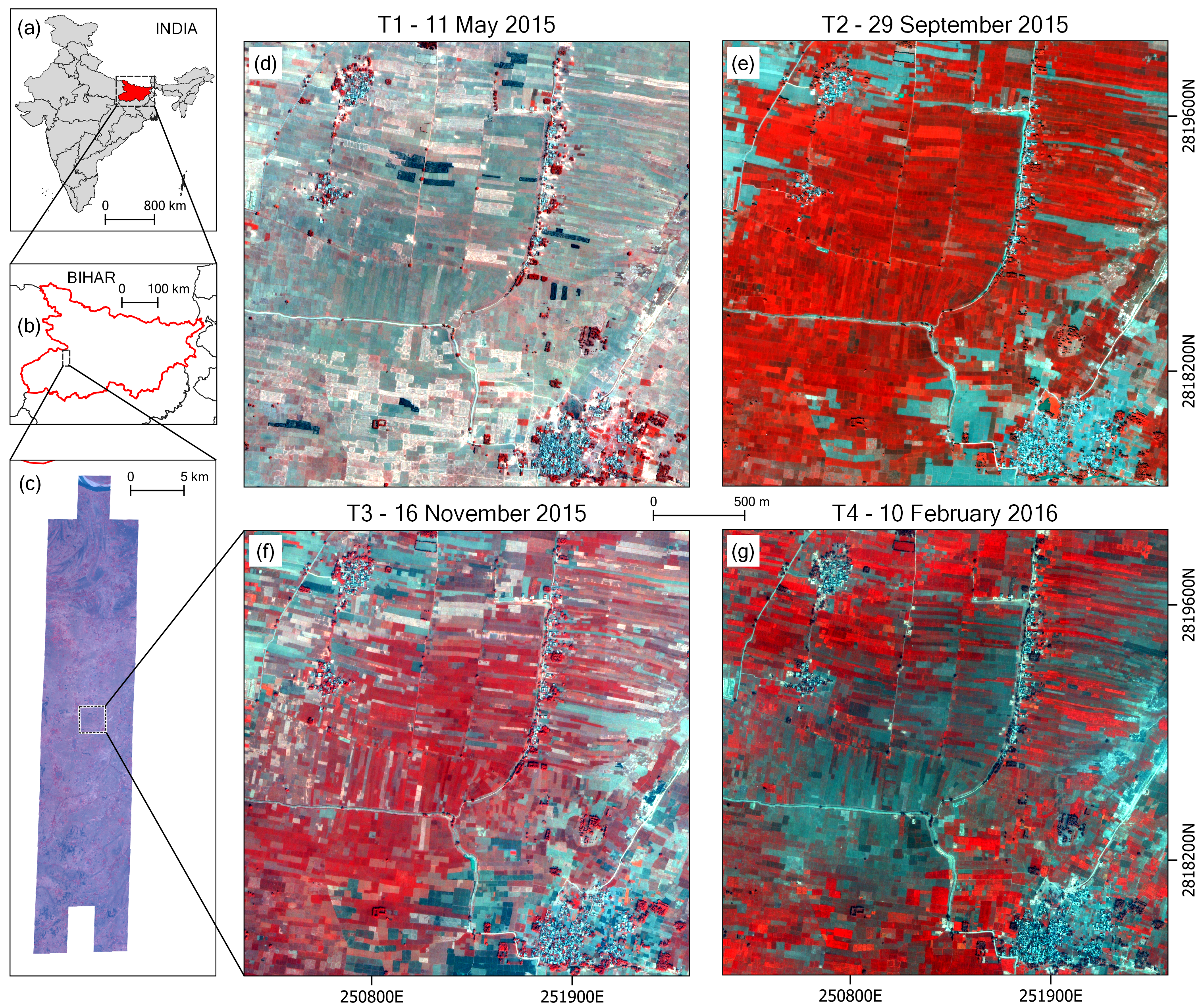}
  \caption{SkySat 2 m resolution satellite images for Bihar, India for dates, (d) 11 May 2015, (e) 29 September 2015, (f) 16 November 2015 and (g) 10 Feb 2016.}
  \label{fig:keymap}
\end{figure*}

\subsection{Ground truth data}
Due to the lack of publicly available field boundaries for accuracy assessment, we manually digitized agricultural field boundaries for our study area. Our reference data consisted of 8,176 polygons across a 6 sq.km. area (Figure \ref{fig:ground_truth}).

\subsection{Accuracy assessment}
We use the Intersection Over Union (IoU) metric for assessing accuracy (see equation \ref{eq:iou}).

\begin{equation}
    \text{IoU} = \frac{\text{intersection area}}{\text{union area}}
    \label{eq:iou}
\end{equation}

Image classification tasks are often evaluated using metrics like confusion matrix, precision, recall, and F1-Scores. However, since our objective is to create an analysis-ready field boundary layer in vector format, we assess accuracy using two metrics: detection accuracy and delineation accuracy \citep{mei2022}. Detection accuracy measures the percentage of correctly identified fields, calculated as the proportion of predicted fields with an IoU value greater than 0.5 compared to the bounding boxes of ground truth polygons. Delineation accuracy on the other hand, evaluates how well the detected fields are delineated, which is calculated as the mean IoU value between the predicted and reference field boundaries (not their bounding boxes).

We perform all processing, including executing SAM, on an Intel(R) Core(TM) i7-12700K (3.60 GHz, 12 cores), 32 GB RAM, Windows 11 OS machine using the SAMGeo Python package \citep{samgeo2023}. The following sections elaborate on the workflow used in this study.

\subsection{Different checkpoints of SAM}
We obtained the three checkpoints of SAM (ViT-b, ViT-h, and ViT-l) available on the GitHub repository associated with \cite{sam2023} to investigate whether the prediction accuracy of any of the three checkpoints outperforms the others. Additionally, we investigate if combining predictions from these checkpoints improves overall accuracy. To achieve this, we merge the predicted polygons into a single layer and compare its accuracy with individual checkpoints.

\subsection{Multi-temporal satellite data}
\label{methods:different_dates_image}
Previous studies have used images from different dates for mapping agricultural fields in various parts of the world \citep{wagner_extracting_2020}. The spectral reflectance of agricultural fields changes substantially during the growth cycle, suggesting that different dates might better capture field edges depending on the crop. Our investigation focuses on smallholder farm settings and utilizes an FM, providing a unique setting to investigate the benefit of using satellite images from different dates. We assess the potential of each image to determine which stages of the crop growth cycle (before plantation, early stage, midway, or mature) impact boundary delineation. This approach allows us to systematically investigate if fusing predictions from multiple dates throughout the crop growth cycle improves field boundary delineation performance.

We identified four cloud-free satellite images taken before and during the 2015-16 winter crop cycle for an area in Bihar, India. These images were acquired on 11 May 2015, 29 September 2015, 16 November 2015, and 10 February 2016, referred to as T1, T2, T3, and T4, respectively, for the remainder of this paper (Figure \ref{fig:keymap}). Bihar is an ideal case for this investigation since it largely consists of small agricultural fields ($<$0.6 ha), and solving the problem in this challenging setting will make it easier to apply to other, comparatively easier settings. The area selected for this work is a square land parcel of 3072 by 3072 pixels (0.8 m cell size); we use this size to generate evenly sized square image chips for the model (details in Section \ref{image_size_method}).

\subsection{Pansharpening \& radiometric correction}
SkySat multispectral data on GEE are available with Blue, Green, Red, and NIR bands at 2 m spatial resolution, and an additional panchromatic band at 0.8 m spatial resolution \citep{planetskysat}. We use the Blue, Green, and NIR bands to generate 3-band false color composite images. We convert the images to byte type by normalizing the values using the 2\textsuperscript{nd} and 98\textsuperscript{th} percentiles for each band and scaling them between 0 and 255, which is crucial since SAM (V1.0) can only be run on byte type images. We use the 2\textsuperscript{nd} and 98\textsuperscript{th} percentile values instead of the minimum and maximum values to avoid outliers.

In addition, we pansharpened the Blue, Green, and NIR bands to 0.8 m using the panchromatic band (equation \ref{eq:pansharpening}).

\begin{equation}
    \text{Pansharpened} = \text{weight} \times \text{band}
    \label{eq:pansharpening}
\end{equation}

Where \text{band} refers to either the Blue, Green, or NIR band; and the pansharpening weight was derived using equation \ref{eq:panweight}.

\begin{equation}
    \text{weight} = \frac{\text{panchromatic}}{\text{blue} + \text{green} + \text{NIR}}
    \label{eq:panweight}
\end{equation}

These preprocessing steps are standard and widely used in the literature \citep{mei2022, Peressutti2023}.

\subsection{Geo-rectification}
Although the SkySat scenes are available as orthorectified products on GEE, we found that the selected SkySat images had poor co-registration (i.e., they were not aligned in 2D space). Due to the lack of an ortho-rectification kit, we manually aligned all four scenes using the \emph{Raster Georeferencer} tool in QGIS LTR 3.22 software \citep{QGIS_software}, using T1 as the reference and 20 tie points manually collected for each of the time periods T2, T3, and T4.

\subsection{Edge enhancement of input images}
In the field of machine learning, feature engineering is a powerful way to transform data and improve model accuracy. However, in remote sensing, there is a lack of consensus on whether feature engineering techniques add value to the processing workflow. For segmenting objects from remote sensing images, edge enhancement methods (increasing image contrast to make linear features prominent) may improve segmentation model performance.

We enhance the edges of our input data and compare the accuracy values with the original (non-edge-enhanced) image output. To determine if the enhanced and original images detect the same or different fields, we combine the predictions from both image types and assess the overall accuracy.

We enhance our pansharpened images using the following equation \citep{mei2022, polesel2000}:

\begin{equation}
    \text{Enh img} = \text{Img} + (\text{Img} - \text{GB img}) \times \text{WF}
\end{equation}

Where \(\text{Enh img}\) refers to the enhanced image, \(\text{Img}\) is the input image, \(\text{GB img}\) is the Gaussian blurred image, and \(\text{WF}\) is the weighting factor.

The Gaussian blurred image is generated using a smoothing radius of 11 pixels and a sigma value (standard deviation of the Gaussian function) of 10 pixels. Upon visually investigating the edge-enhanced outputs from different weighting factor values, we find that a weighting factor of 2 works best, making boundaries clearer without adding excessive noise to the image.

\subsection{Investigating different sizes of input image}
\label{image_size_method}
Since SAM can be run on images of any size, we systematically investigate the accuracy associated with different input image sizes. We clip the images for all four time periods into smaller tiles of four sizes: 256 x 256, 512 x 512, 768 x 768, and 1024 x 1024 pixels. We avoid sizes smaller than 256 or larger than 1024 due to the number of field boundaries being too few or too many, respectively. To test whether the same or different fields are detected in each image size, we combine the predicted polygons into one layer and assess its accuracy.

Clipping the input satellite images into different sizes results in partitioned polygons at the edges of images, which we correct by developing a GIS workflow that merges these polygons in adjacent layers. This is achieved by identifying the polygons in adjacent layers that overlap each other's boundary. If the overlap on the overlapping edge of both polygons is greater than 85\%, we merge the polygons (see Figure \ref{fig:adjacent_poly}).

\subsection{Different levels of accuracy assessment}
\label{sec:accuracy_setup}
\begin{figure}
\centering
\includegraphics[width=0.45\textwidth]{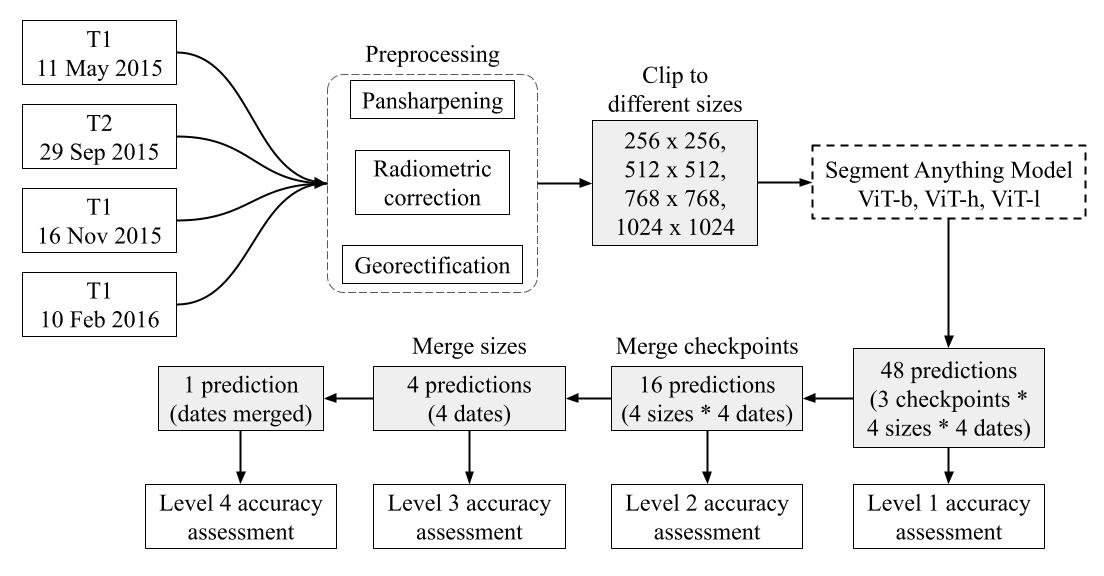}
\caption{Schematic diagram of the different levels of accuracy assessment for predicted layers.}
\label{fig:flowchart}
\end{figure}

We evaluate the accuracy of each of the 48 predicted layers individually and by combining them step by step, referring to these steps as different levels of accuracy assessment (Figure \ref{fig:flowchart}). Level 1 assesses the accuracy of all 48 layers directly from the model. In Level 2, we combine the predicted vector layers from all the SAM checkpoints to obtain 16 predicted layers (four different image sizes × four different dates). Level 3 involves combining predictions from different sizes to narrow down to four predicted layers, each representing a different date. In Level 4, we combine all the dates to generate the final predicted layer. We perform these four levels of accuracy assessment separately for both the original and edge-enhanced images.

\section{Results and discussion}
\subsection{Performance across SAM checkpoints}
\begin{figure}
\centering
\includegraphics[width=0.45\textwidth]{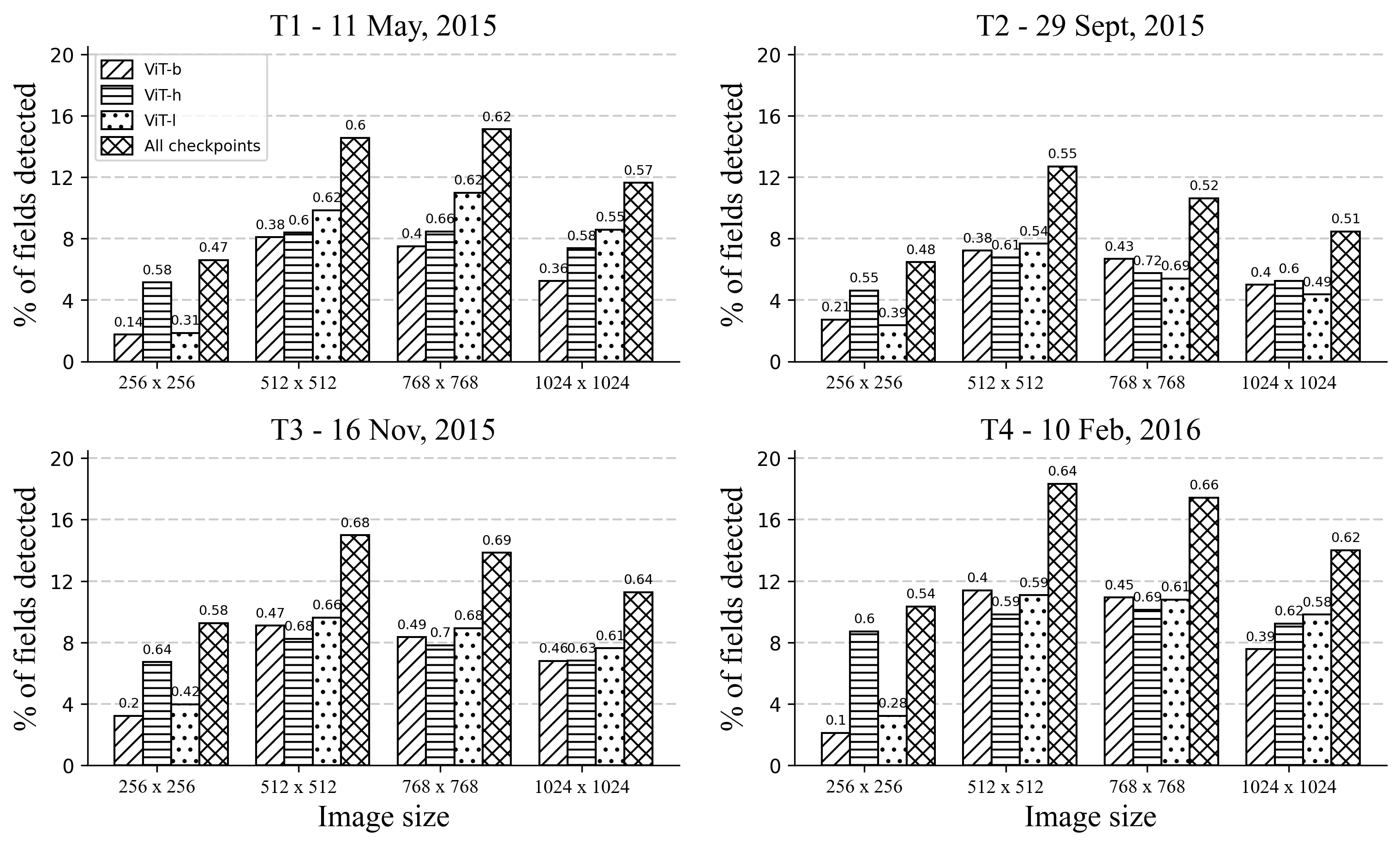}
\caption{Detection and delineation accuracy at level 1 (raw predictions) and level 2 (after combining predictions from different checkpoints). Different bars represent different checkpoints. The Y axis represents detection accuracy, and the X axis shows bars grouped by input image chip size. The text on top of each bar shows the mean IoU value for that group.}
\label{fig:checkpoints_combined}
\end{figure}

First, we examine whether different SAM checkpoints (ViT-b, ViT-h, and ViT-l) produce consistent results or if one performs better. Figure \ref{fig:checkpoints_combined} illustrates the first level of accuracy assessment metrics for 48 predicted layers (three checkpoints × four image sizes × four dates). The detection accuracy (percentage of fields detected) varies across different checkpoints, with none systematically outperforming the other two.

By combining the predictions of all three checkpoints, the number of layers reduce from 48 to 16 (four sizes × four dates). Combining layers from different checkpoints results in a slight improvement in detection accuracy compared to each individual checkpoint across all image sizes and dates (see Figure \ref{fig:checkpoints_combined}). For instance, the three checkpoints in the T4 512 x 512 set are in the 14-15\% detection range, but their combined output reaches 18\% detection accuracy. This indicates that while all three checkpoints largely detect the same fields, some fields are unique to each checkpoint. Moreover, the combined checkpoint in the case of T4 512 x 512 also exhibits higher delineation accuracy (IoU value 0.69) than its constituent checkpoints (0.61 - 0.64). This suggests that the shape of the fields detected redundantly across three checkpoints is captured differently by each checkpoint. Therefore, combining predictions from different checkpoints can improve overall prediction accuracy, with a maximum detection accuracy of 18\% in this case.

\subsection{Impact of Input Image Size}
\label{discussion:image_sizes}
\begin{figure}
\centering
\includegraphics[width=0.45\textwidth]{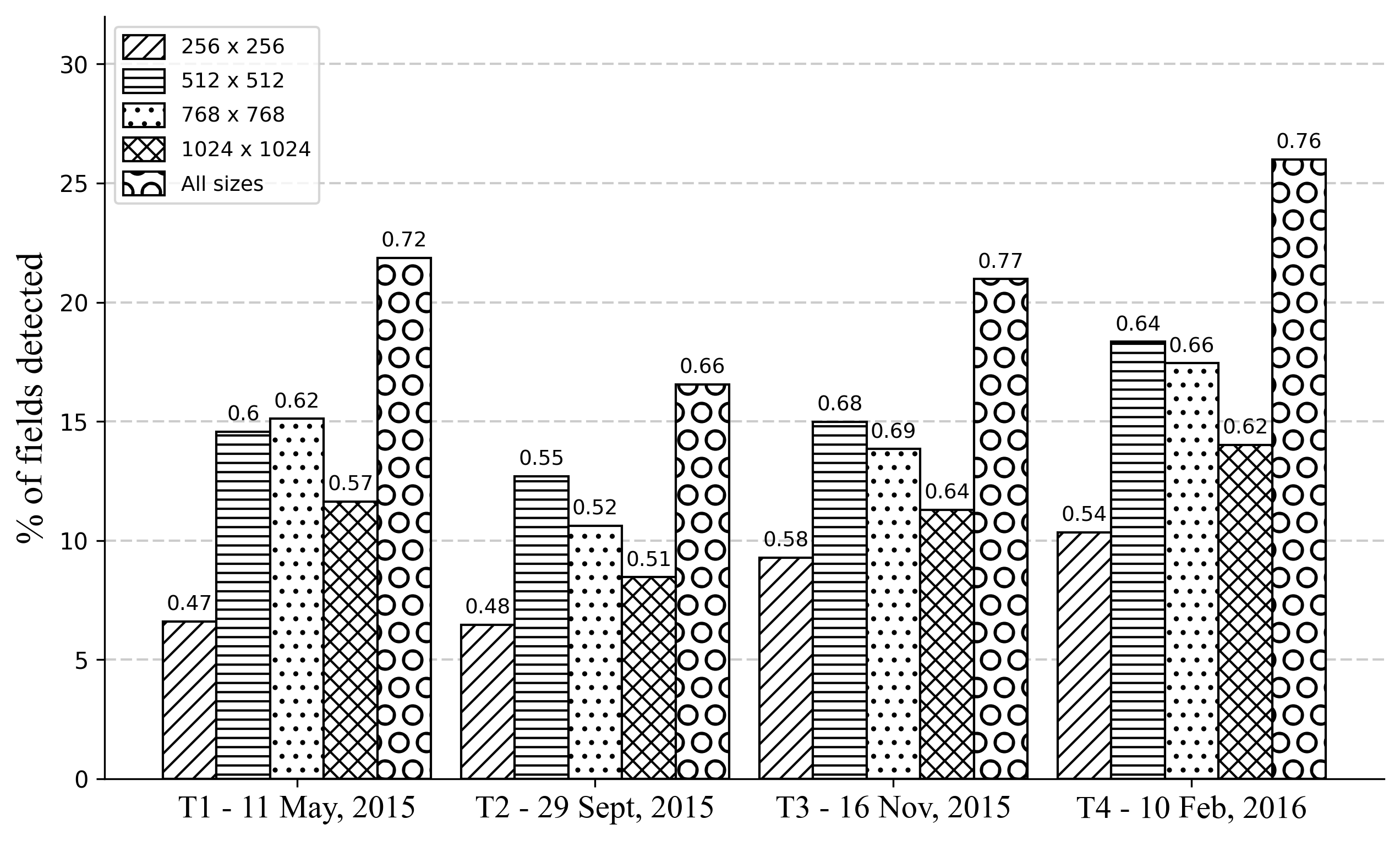}
\caption{Accuracy metrics at level 2 (different image dates after combining checkpoints) and level 3 (combining image sizes).}
\label{fig:sizes_combined}
\end{figure}

Next, we investigate the impact of input image size on segmentation accuracy. Figure \ref{fig:sizes_combined} illustrates the detection and delineation accuracy values of predictions from different image sizes. Detection accuracy is systematically higher for mid-range image sizes (512 x 512 and 768 x 768) and lower for the smallest (256 x 256) and largest (1024 x 1024) images. This could be because, below a certain image size, SAM fails to identify boundaries and sometimes predicts the entire image chip as one big parcel (see Appendix Figure \ref{fig:block_prediction}). Conversely, when the input image size is too large (1024 x 1024), SAM fails to identify smaller field boundaries, resulting in a drop in detection accuracy. Therefore, for this 2 m resolution data pansharpened to 0.8 m, the 512 x 512 and 768 x 768 image sizes yield the best results for smallholder farm boundary delineation.

Combining the predictions from different image sizes and assessing the accuracy consistently results in a significant rise in detection accuracy across all dates (Figure \ref{fig:sizes_combined}). For instance, the maximum detection accuracy among all four image sizes in T4 is 18\%, and the accuracy of the size-combined version increases to 26\%. Interestingly, while detection accuracy increases, delineation accuracy remains consistent between 0.67 and 0.71.

\subsection{Multi-temporal Satellite Data}
\begin{figure}
\centering
\includegraphics[width=0.4\textwidth]{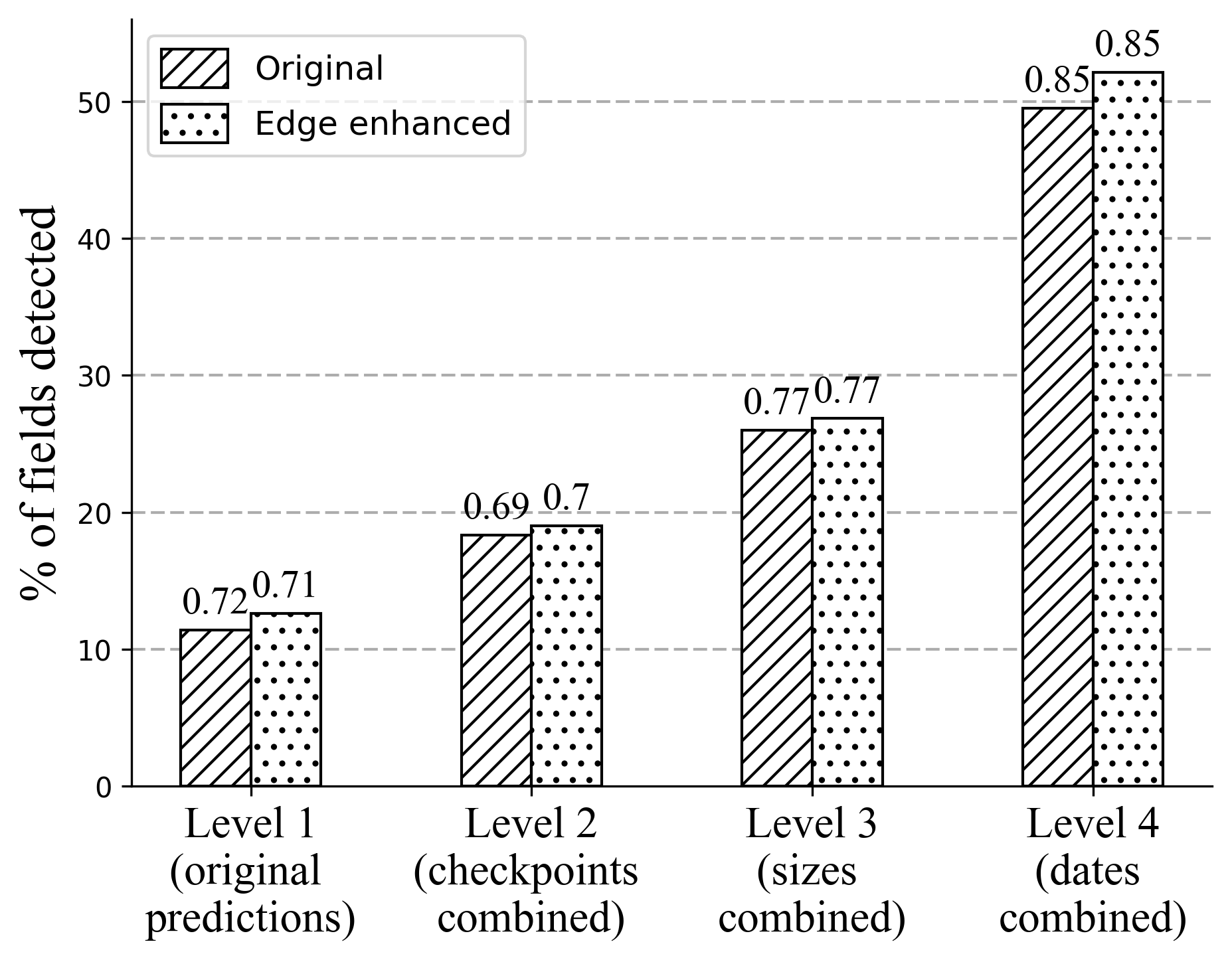}
\caption{Accuracy metrics for original and edge enhanced images at different level of accuracy assessment.}
\label{fig:everything_combined}
\end{figure}

We then explore whether using satellite images from different dates impacts the predictions. Figure \ref{fig:everything_combined} shows that combining predictions from satellite images taken on different dates significantly improves the detection percentage—from 26\% to 50\%. This improvement can be attributed to some fields being better identified at early stages of the crop growth cycle, while others are delineated better as the crop grows. For instance, if the same crop type is planted in adjacent fields, the likelihood of crops spilling over and covering the boundaries between two fields is higher as the crop grows. Conversely, if two different crop types are planted in adjacent fields, it becomes easier for the model to delineate the two fields towards the later stage of the crop growth cycle.

\subsection{Enhancing Images}
Finally, we examine whether enhancing the edges of the input images improves segmentation accuracy. All the metrics shown for the original images in Figures \ref{fig:checkpoints_combined} and \ref{fig:sizes_combined} exhibit a similar pattern for edge-enhanced images (Figures \ref{fig:enhanced_dates} and \ref{fig:enhanced_sizes}). Figure \ref{fig:everything_combined} presents a summary of the improvement in accuracy metrics for edge-enhanced images at different levels of accuracy assessment. The non-edge-enhanced images detect roughly 50\% of the fields, with the edge-enhanced version capturing 53\%. To empirically investigate redundancy between these two, we combined both layers and found an increase in detection accuracy to roughly 58\% (see Table \ref{tab:performance}).

\begin{table}[h]
    \centering
    \small
    \begin{tabular}{lccccc}
    \hline
     & Detection (\%) & IoU & Precision & Recall & F1 \\
    \hline
    Raw & 50 & 0.71 & 0.79 & 0.89 & 0.82 \\
    Enhanced & 53.74 & 0.71 & 0.79 & 0.88 & 0.82 \\
    Combined & 57.89 & 0.71 & 0.79 & 0.88 & 0.82 \\
    \hline
    \end{tabular}
    \caption{Performance Metrics for original and edge enhanced images, and their combined version.}
    \label{tab:performance}
\end{table}

This indicates that while edge enhancement alone does not drastically improve accuracy, the synergistic use of both original and enhanced images can achieve maximum possible accuracy—in this case, from 50\% to 58\%. This finding deviates from previous work that found no improvement in accuracy when using edge-enhanced images \citep{mei2022}. This difference could likely be due to variations in satellite data resolution and the models used. Additionally, Table \ref{tab:performance} lists the Precision, Recall, and F1-Score values for the matched polygons, showing that the predicted polygons are well delineated.

\subsection{Smallholder polygons}
\begin{figure}
\centering
\includegraphics[width=0.49\textwidth]{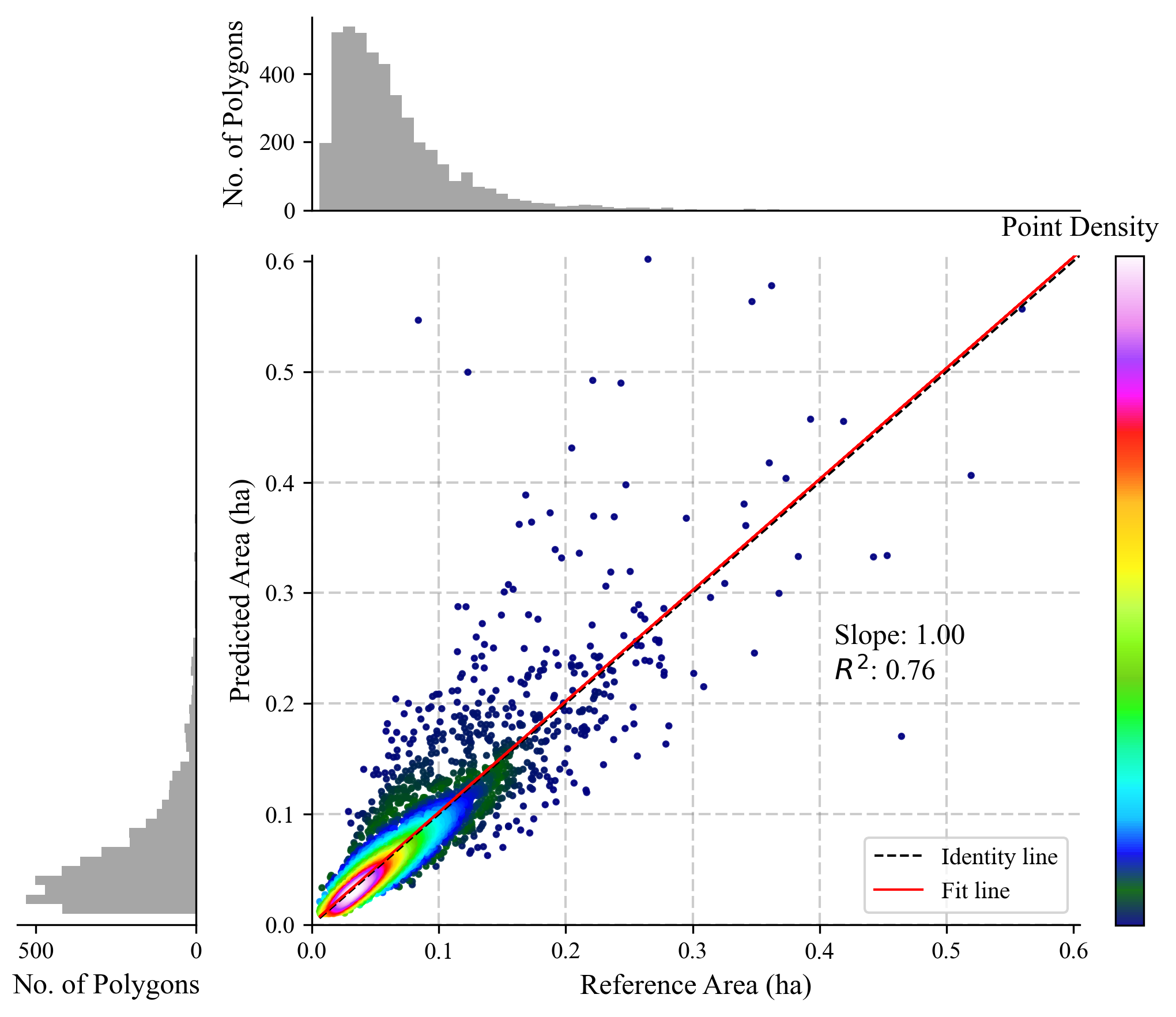}
\caption{Histogram and scatter plot of area of actual and predicted field boundaries.}
\label{fig:scatterplot}
\end{figure}

Figure \ref{fig:scatterplot} presents histograms of the predicted and actual polygon areas, along with a scatter plot comparing their areas. Most predictions are close to the identity line, with some outliers indicating both over- and under-prediction. Notably, over 70 percent of the detected polygons have an area below 0.1 ha, which is much smaller than the expected smallholder farm size of 0.6 ha. Despite this, our proposed workflow successfully captures these smallholder farm boundaries using 2 m resolution satellite data pansharpened to 0.8 m data without any additional model training.

Our results indicate that using all three SAM checkpoints, which have largely overlapping predictions, can improve detection accuracy when combined. Additionally, input image sizes that are neither too small nor too large prove to be more effective. Similar to combining checkpoints, combining predictions from different image sizes also enhances detection accuracy. However, the most significant improvement in detection accuracy is observed on combining predictions from different dates. Notably, merging predictions from both the original and edge-enhanced images improves the detection accuracy of agricultural fields.

Overall, the accuracy of SAM in detecting agricultural field boundaries is promising, particularly given the challenging conditions of using 2 m resolution data without additional training labels. For comparison, \cite{mei2022} achieved an average IoU value of 0.85 and F1 Scores greater than 0.72 using very-high-resolution (0.5 m) imagery and Mask R-CNN for mapping smallholder fields in India. While SAM’s performance with 2 m resolution data may not surpass these results, it remains competitive, especially considering the lack of labeled training data and the lower resolution.

\subsection{Future Work}
\label{discussion:future_work}
While SAM can detect approximately 58\% of agricultural fields without training labels, identifying correctly detected fields still relies on ground truth labels. These "good" delineations were identified using an IoU value threshold of 0.5 between the bounding box of the predicted field and the ground truth data. Future work could explore using the promptable functionality of SAM to address this limitation.

Moreover, our current setup does not recognize aggregated polygons (see \ref{fig:aggregate_prediction}) as valid predictions because an aggregated polygon would not have an IoU value of 0.5 with any of the constituting polygons in the reference layer. We set our accuracy assessment this way to get the number of correctly detected polygons instead of the total area correctly delineated. Our rigorous evaluation approach of not considering aggregated polygons may underestimate our workflow's accuracy. Calculating layer-level metrics instead of individual polygon-level metrics, which include all aggregated polygon predictions, would likely result in higher accuracy compared to our current approach that discards aggregated polygon predictions. However, this work focuses on investigating ways to maximize SAM's performance for agricultural boundary detection and therefore distances itself from dealing with aggregated polygon predictions.

Additionally, co-registration issues in high-resolution images can be a bottleneck when using images from different dates. Future efforts should ensure high-quality co-registration to scale up the findings of this study effectively.

Although detecting 58\% of smallholder farm boundaries is a significant achievement, further work is needed to enhance SAM's detection capabilities. Utilizing satellite images from more dates, as shown in this work, significantly improves detection accuracy and may be used to further enhance performance.

\section{Conclusion}
This paper contributes to the rapidly evolving domain of using foundation models and GeoAI for environmental applications, specifically investigating SAM for mapping smallholder agricultural farm boundaries in data-scarce regions. Using freely available SkySat satellite data from different dates, our findings show that combining different SAM checkpoints slightly improves accuracy. More notably, combining predictions from images of various input sizes improved detection accuracy, with the greatest improvement observed when using images from different dates. Edge enhancement of the input satellite images further boosts overall detection accuracy, indicating the value of feature engineering techniques in this context. By integrating these model variations, we created a comprehensive agricultural field boundary layer, detecting 58\% of fields without additional training.

This work serves as a proof of concept for using SAM as a valuable tool for field boundary mapping in data-scarce environments, highlighting its capability to perform without any training while also expediting training data generation for other models.

Moreover, this work aligns with the principles of Geospatial Explainable Artificial Intelligence (GeoXAI) by enhancing the transparency and interpretability of AI-driven agricultural mapping. The combined use of different model checkpoints, image sizes, and dates, along with feature engineering, not only improves accuracy but also provides insights into the model’s decision-making process, fostering trust and reliability in AI applications for geospatial analysis.

\section*{Acknowledgements}
Pratyush Tripathy acknowledges support from the Benioff Scholars Program in Applied Environmental Science and the Schmidt Family Foundation Environmental Solutions Fellowship at the University of California, Santa Barbara.

\bibliography{main}

\clearpage

\onecolumn
\appendix
\counterwithin{figure}{section}

\section{Data creation and pre-processing}
\label{appendix_data_creation}

\begin{figure}[H]
\centering
\includegraphics[width=0.45\textwidth]{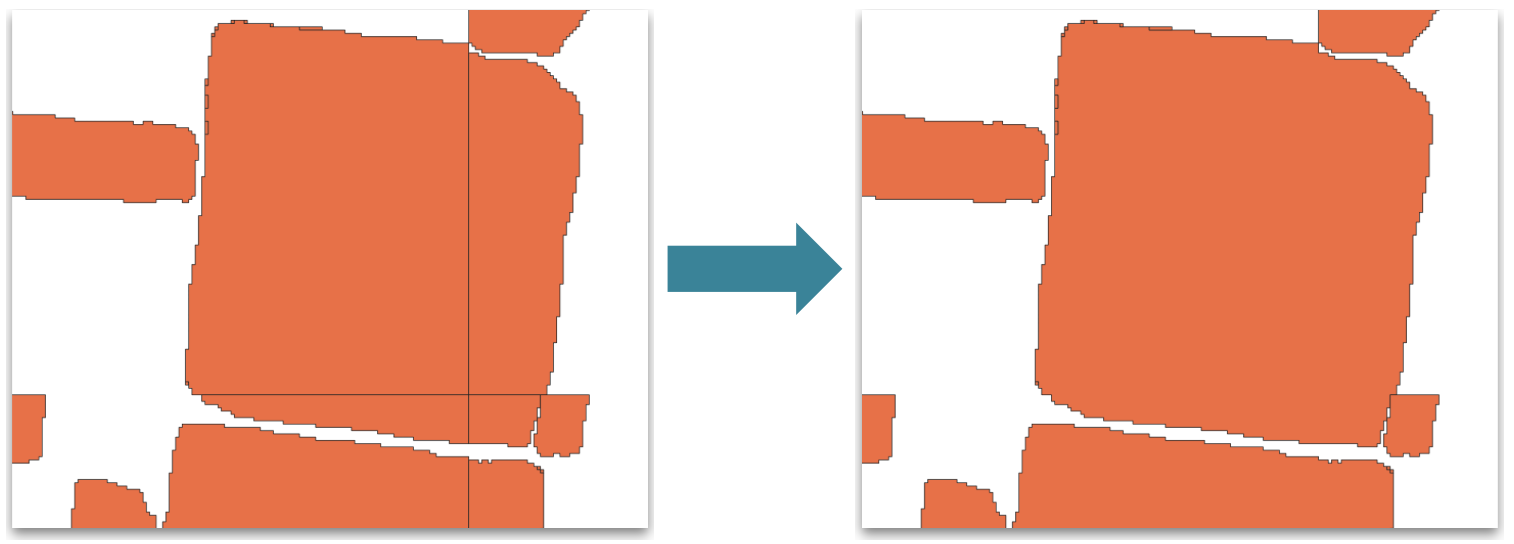}
\caption{Polygons in the adjacent tiles before merging and after merging. Workflow described in Section \ref{image_size_method}.}
\label{fig:adjacent_poly}
\end{figure}

\begin{figure}[H]
\centering
\includegraphics[width=0.45\textwidth]{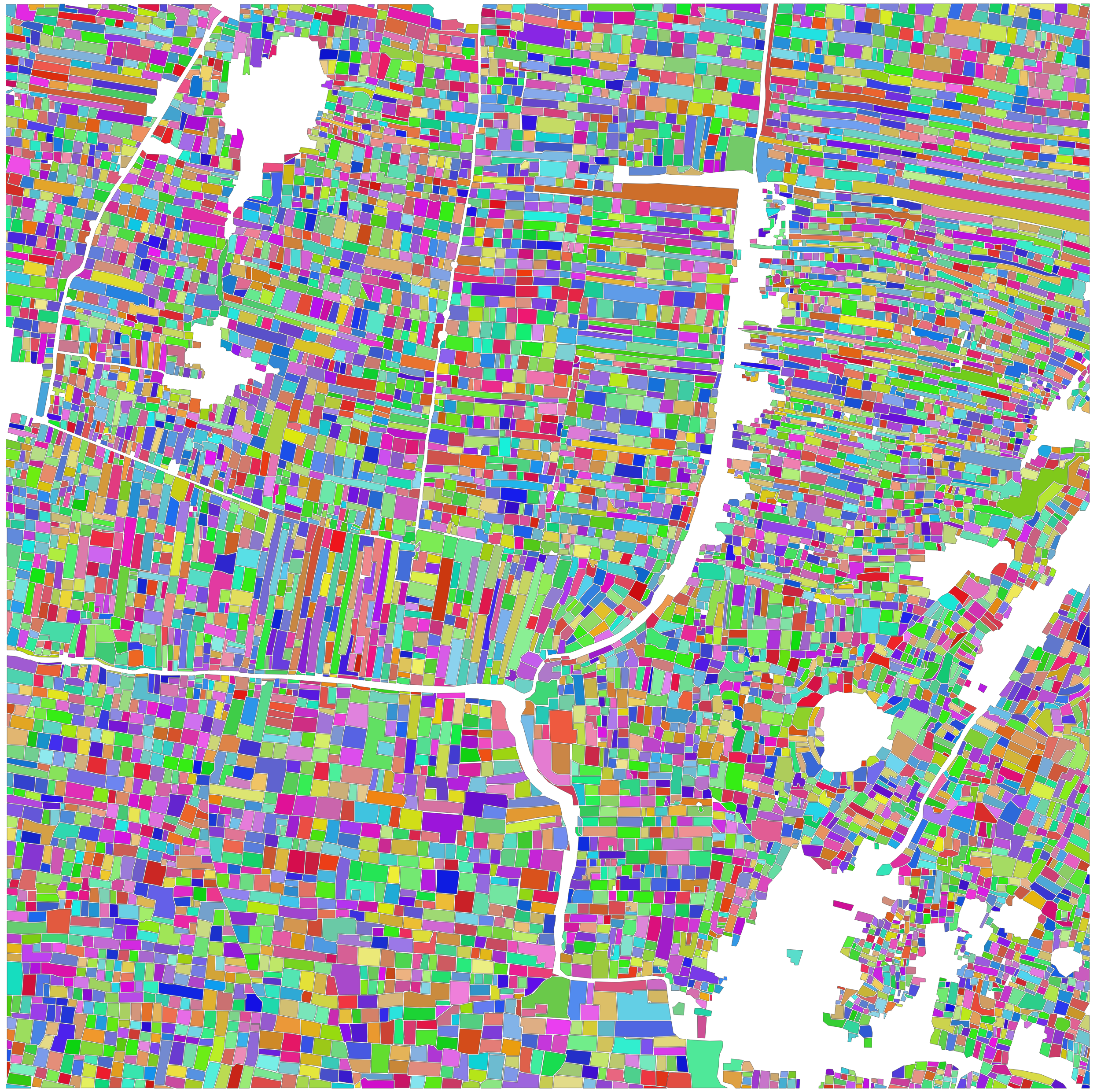}
\caption{Manually generated ground truth agricultural field boundary polygons.}
\label{fig:ground_truth}
\end{figure}

\begin{figure}[H]
\centering
\includegraphics[width=0.45\textwidth]{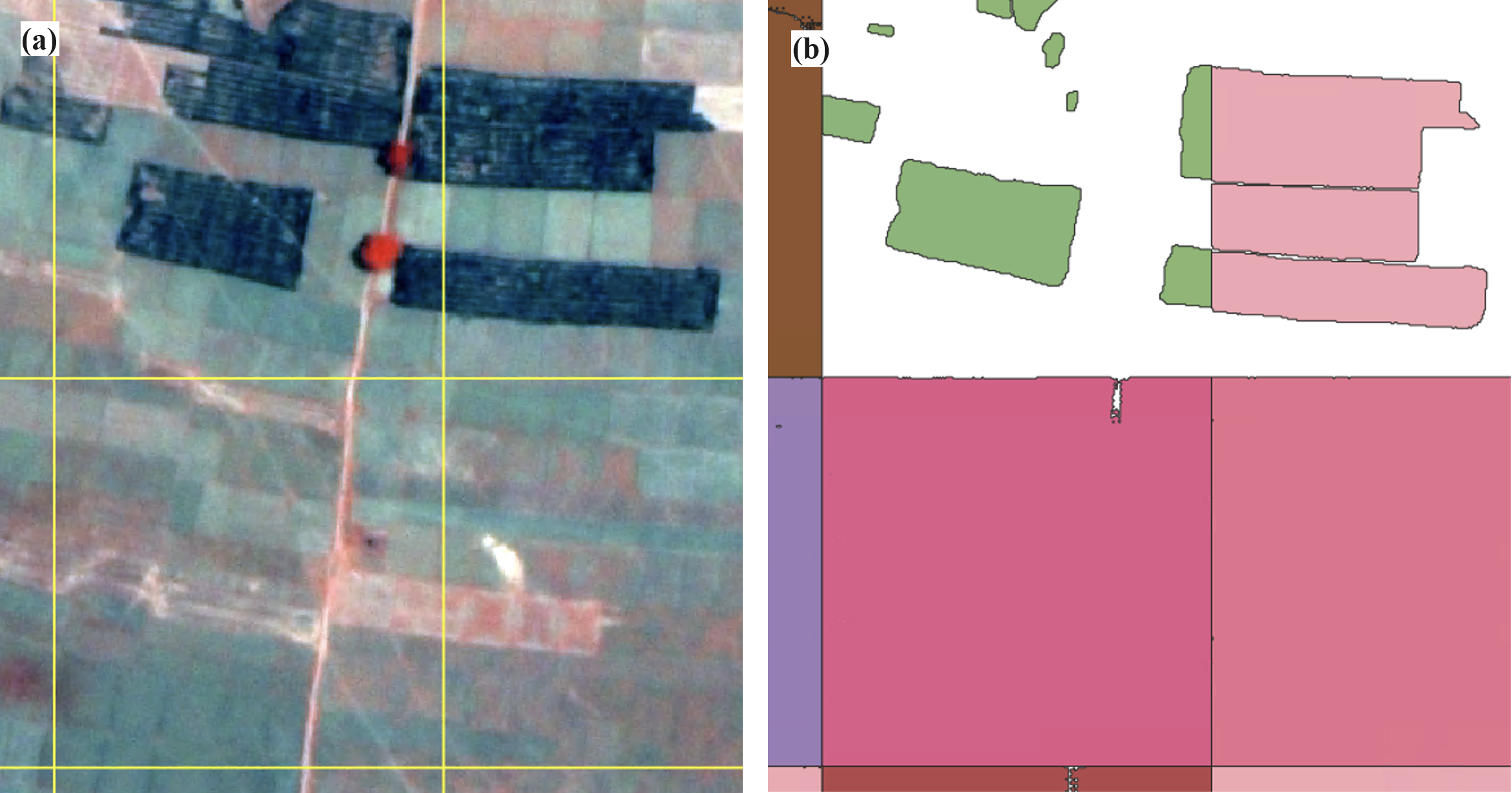}
\caption{Image showing the entire image chip (size 256 by 256) being predicted as one large polygon. More discussion in Section \ref{discussion:image_sizes}}
\label{fig:block_prediction}
\end{figure}

\section{Predictions}
\label{appendix_predicitons}

\begin{figure}[H]
\centering
\includegraphics[width=0.45\textwidth]{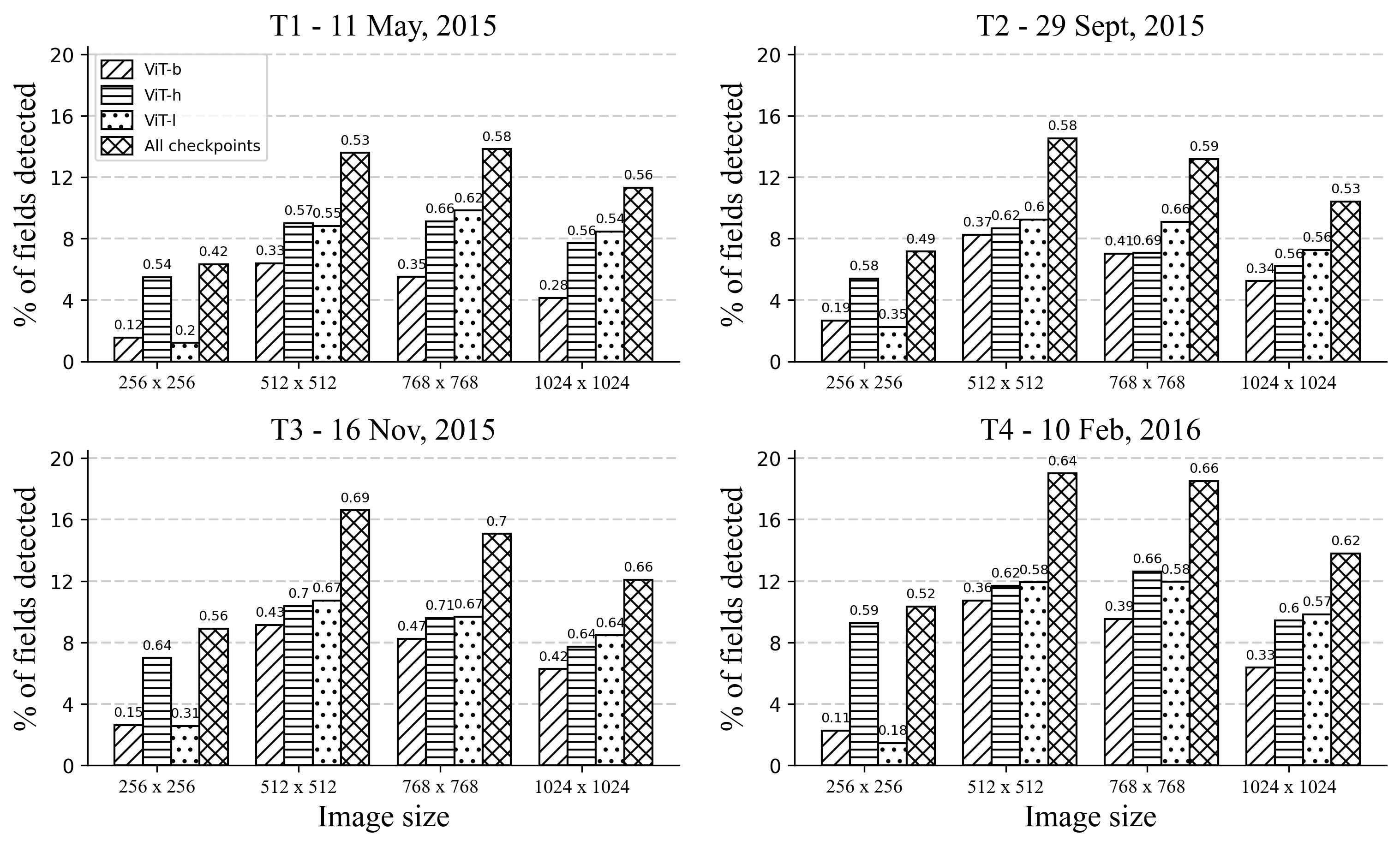}
\caption{Accuracy metrics of different image sizes and their combined version for edge-enhanced input images.}
\label{fig:enhanced_sizes}
\end{figure}

\begin{figure}[H]
\centering
\includegraphics[width=0.45\textwidth]{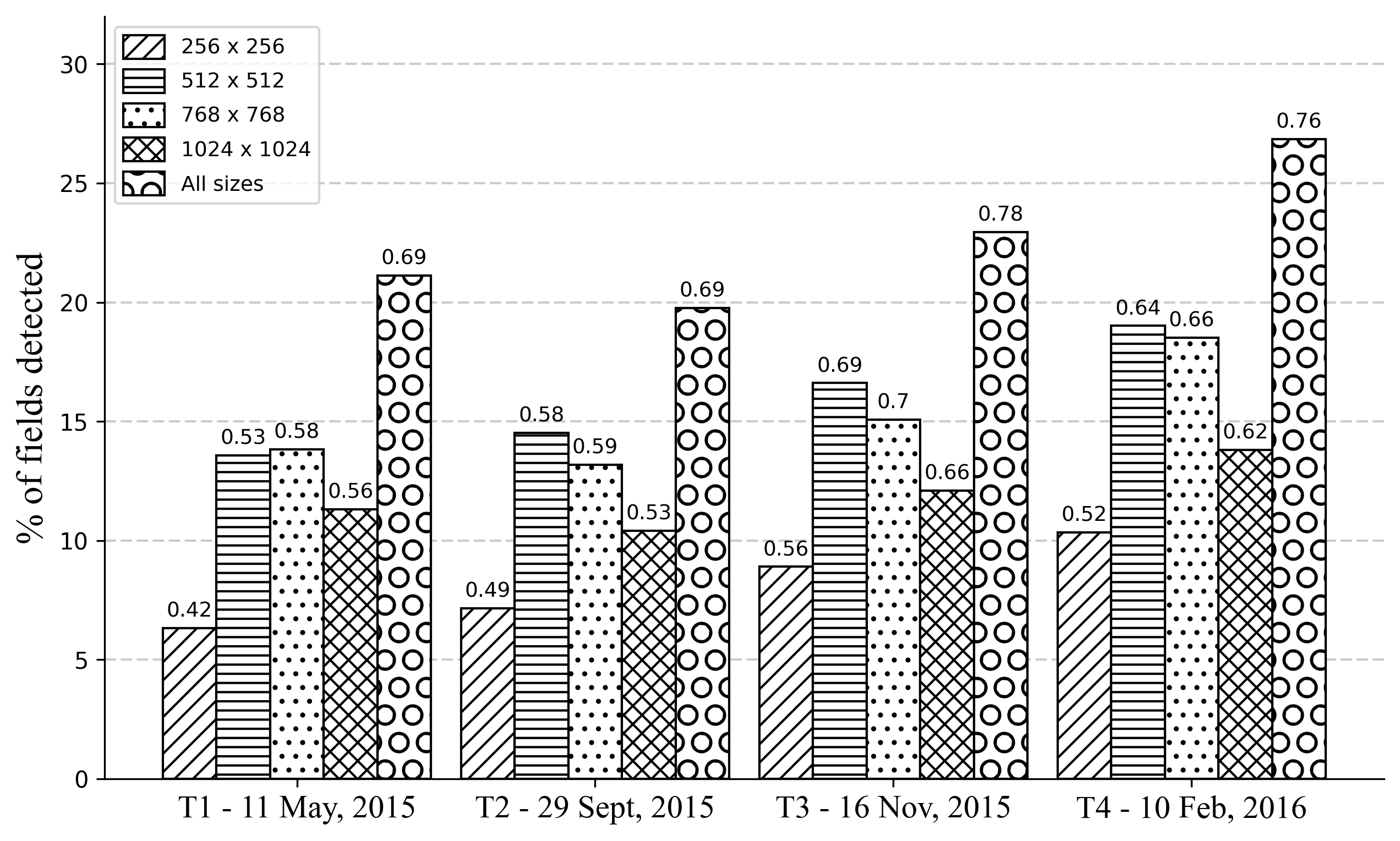}
\caption{Accuracy metrics of different image dates and their combined version for edge-enhanced input images.}
\label{fig:enhanced_dates}
\end{figure}

\begin{figure}[H]
\centering
\includegraphics[width=0.45\textwidth]{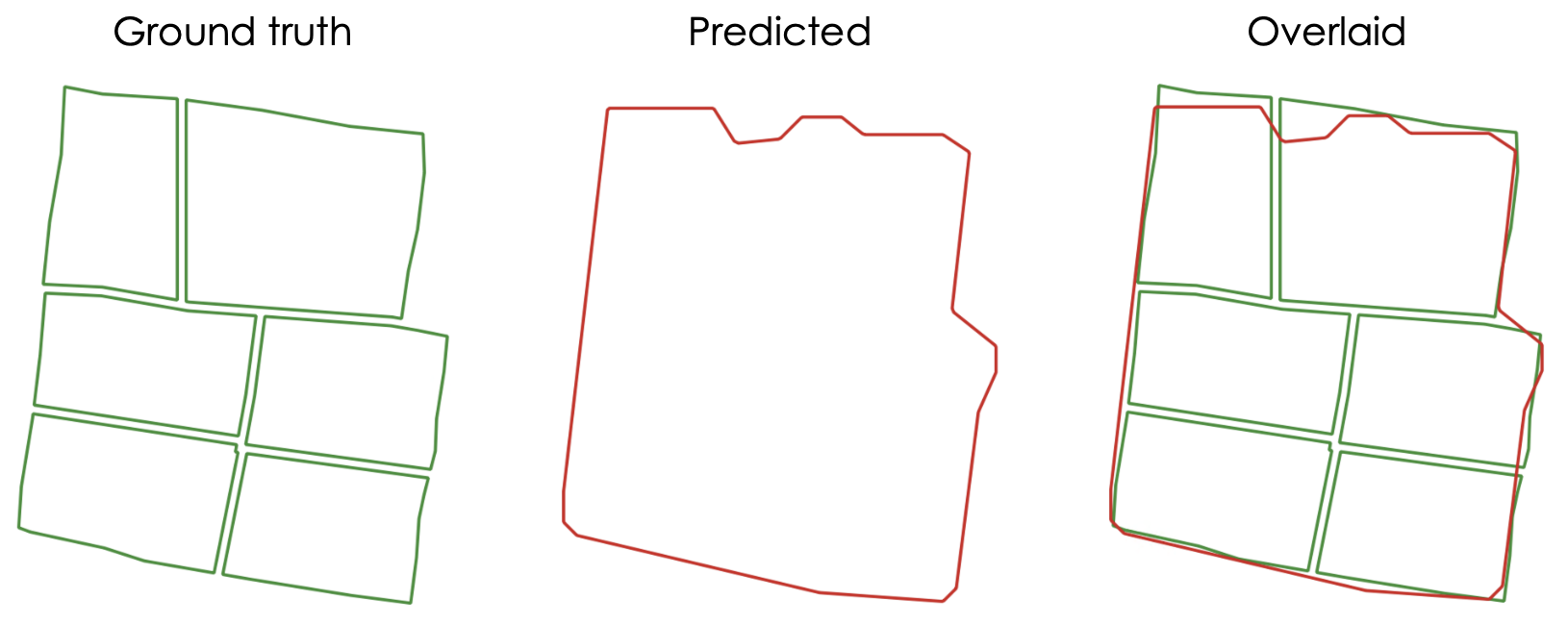}
\caption{Figure showing a group of smallholder farms being detected as one single large agriculture field. More discussion on this in Section \ref{discussion:future_work}.}
\label{fig:aggregate_prediction}
\end{figure}

\begin{figure}[H]
\centering
\includegraphics[width=0.9\textwidth]{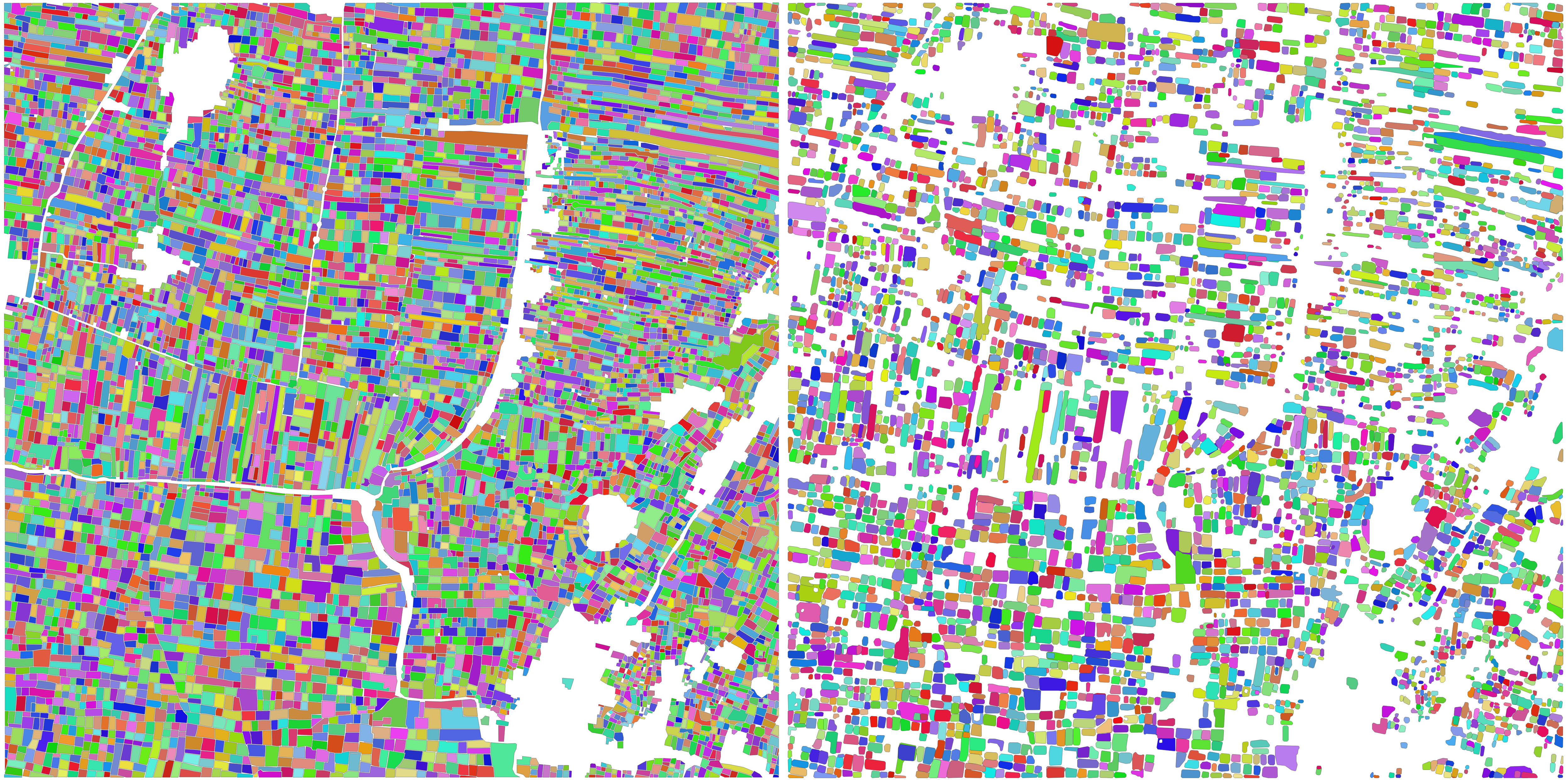}
\caption{(a) ground truth and (b) predicted agricultural field boundaries showing 58\% detected fields.}
\label{fig:predicted_polys}
\end{figure}

\end{document}